\documentclass[runningheads]{llncs}

\usepackage{eccv}

\usepackage{eccvabbrv}
\usepackage{enumitem}   

\usepackage{graphicx}
\usepackage{booktabs}
\usepackage{caption}
\usepackage{multirow}

\usepackage[accsupp]{axessibility}  

\usepackage{hyperref}

\usepackage{orcidlink}

\begin{document}

\title{RegVGGT: Sustainable Visual Geometry Grounding for Streaming via Regulated Memory} 

\titlerunning{RegVGGT}

\author{Hongbo Mao \and
Junjun Jiang\thanks{Corresponding anthor.}\orcidlink{0000-0002-5694-505X} \and
Youyu Chen \and
Jiaxin Zhang \and
Zhemeng Dong \and
Xianming Liu}

\authorrunning{H.~Mao et al.}

\institute{Harbin Institute of Technology, Harbin 150001, China 
\\
\email{jiangjunjun@hit.edu.cn}
\\
\url{https://github.com/amao996/RegVGGT}}

\maketitle

\begin{center}
  \centering
  \includegraphics[width=1.0\linewidth]{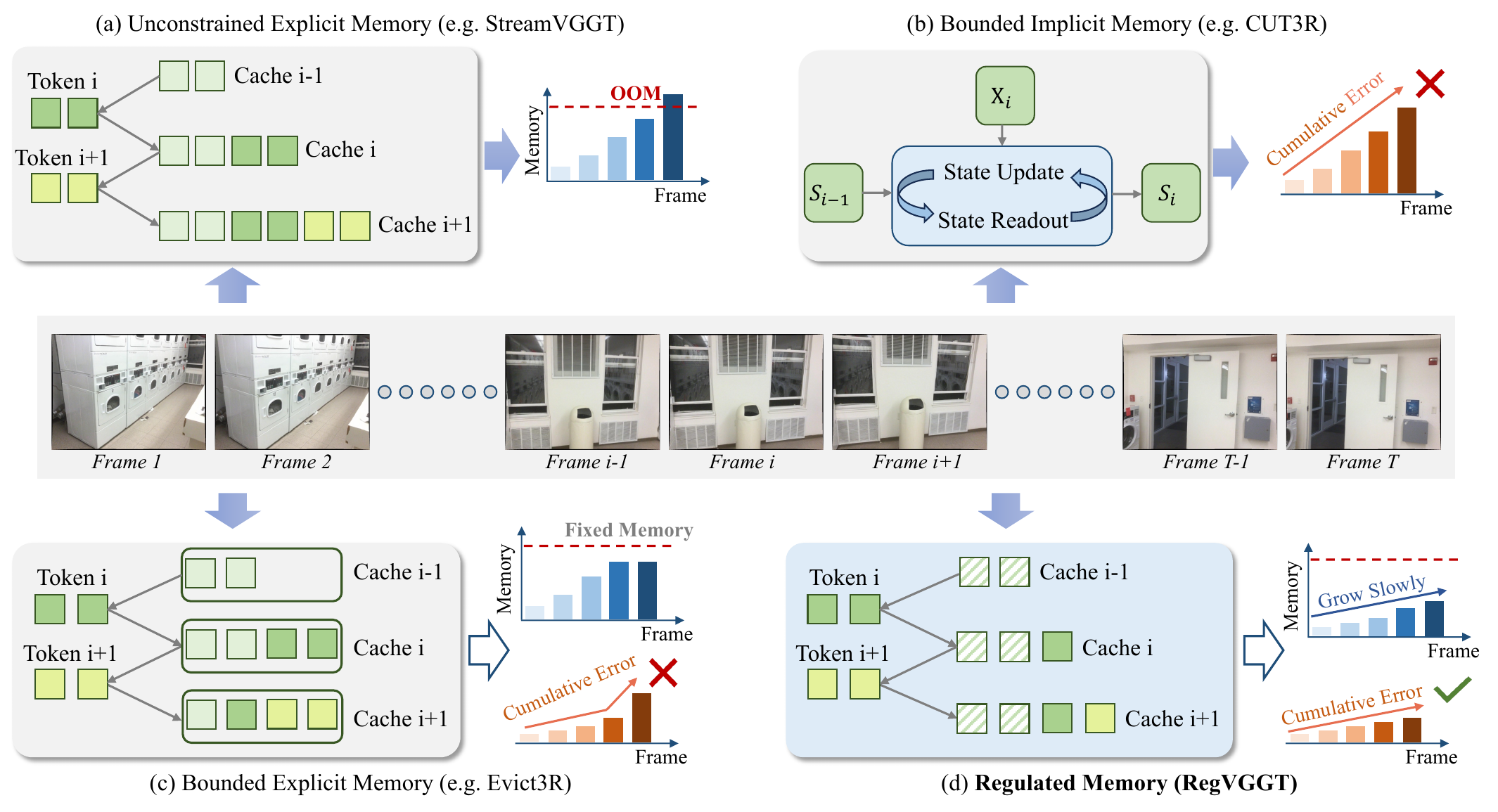}
	\captionof{figure}{\textbf{Paradigm Comparison} between previous streaming reconstruction paradigms and our RegVGGT. 
    (a) Unconstrained explicit memory retains all historical KV pairs, causing memory usage to inflate rapidly. 
    (b) Bounded implicit memory compresses historical information into compact hidden states, resulting in catastrophic forgetting. 
    (c) Bounded explicit memory enforces a fixed memory budget, relying on the implicit prior of scene scales to properly measure the budget.
    (d) Our regulated memory admits at most 1\% of tokens per frame to update the context memory, dramatically suppressing the memory inflation as the stream progresses.}
  \label{fig:teaser}
\end{center}

\begin{abstract}
  3D reconstruction from a lengthy video stream input poses a dilemma for \textbf{f}eed-\textbf{f}orward \textbf{r}econstruction \textbf{m}odels (FFRMs), that a whole-stream inference context cannot be retained under limited GPU memory. 
  Recent studies seek to resolve this problem via a trade-off between the integrity of inference context and GPU memory usage, which either suffer from a rapid memory inflation or degraded context integrity due to artificially capping memory usage. 
  Driven by our key observation that the initial saliency of a token reliably dictates its long-term importance across the stream, we propose \textbf{RegVGGT}, a training-free token regulation method which aggressively regulates the tokens of incoming frames. 
  By admitting at most 1\% of tokens per frame to update the context memory, our method dramatically suppresses memory inflation as the stream progresses. 
  Equipped with a FlashAttention-compatible token saliency estimation scheme, RegVGGT is capable of processing thousands of frames on a consumer-grade GPU with negligible compromise to reconstruction quality. 
  Extensive experiments demonstrate that RegVGGT achieves state-of-the-art performance on long-horizon benchmarks across diverse FFRM prediction tasks, surpassing prior FFRM-based stream reconstruction baselines by a large margin. 
\end{abstract}

\section{Introduction}
\label{sec:intro}
Dense 3D geometry reconstruction from 2D image collections has long been a key topic of 3D computer vision, underpinning a variety of downstream applications such as augmented reality (AR)~\cite{zheng2024gps,hong20243d,lei2025mosca}, embodied robotics~\cite{liu2024robust,wang2024embodiedscan,wu2025embodiedocc}, and autonomous driving~\cite{huang2023tri,huang2024sgaussian,zhou2024drivinggaussian}. 
Building upon decades of foundational research~\cite{wu2013towards,furukawa2009accurate,schonberger2016structure}, recent advances have achieved an end-to-end reconstruction paradigm to predict dense 3D geometry directly from input images in a feed-forward manner~\cite{wang2024dust3r,wang2025vggt,wang2025pi}. 
These feed-forward reconstruction models (FFRMs) are built upon visual transformers~\cite{dosovitskiy2020image}, which embed input images into tokens, perform reconstruction with attention mechanisms~\cite{vaswani2017attention,dao2022flashattention} across the tokens, and decode the tokens into dense 3D geometry such as point maps and camera poses. 

Despite their remarkable achievements, these FFRMs operate in an offline inference regime that forces reprocessing all historical frames whenever a new frame arrives, making it infeasible to process lengthy video streams. 
By implementing the token attention operation using a temporal causal approach to replace the canonical global attention, recent studies~\cite{zhuo2025streaming,lan2025stream3r} adapt FFRMs to process incoming frames one-by-one, namely online inference. 
This largely alleviates the computational overhead for stream reconstruction, because the historical tokens can be directly accessed from memory eliminating the need to recompute them as each new frame arrives.
To further extend the capacity of historical frame context embedded into limited memory, various memory bank strategies have been introduced to maintain historical frame tokens as context implicitly~\cite{wang2025continuous,li2025wint3r,chen2025ttt3r} or explicitly~\cite{mahdi2025evict3r,su2026xstreamvggt,yuan2026infinitevggt} for reconstructing the incoming frames, as shown in \cref{fig:teaser}. 
While these methods succeed in capping memory usage, they suffer from forgetting crucial historical context as the processed frames increase, resulting in degraded reconstruction accuracy. 

In this paper, we propose RegVGGT, a training-free token regulation method designed to adapt a prior FFRM~\cite{zhuo2025streaming} to lengthy stream input, which \textit{merely retains at most 1\% tokens for each incoming frame} as the context memory for future frames. 
Our method stands upon a key observation, that the saliency of each FFRM token to all other tokens across the stream is temporally consistent. 
Thus, the initial saliency of a FFRM token reliably dictates its long-term importance across the stream. 
This allows us to aggressively prune redundant tokens online for each frame without discarding possibly salient tokens for the future frames. 
To further improve the efficiency for temporal causal attention of our method, we introduce a FlashAttention-compatible saliency estimation mechanism that recovers precise attention scores on-the-fly via Log-Sum-Exp (LSE) statistics without materializing the complete attention map. 
Applying our method to a pretrained FFRM using temporal causal attention~\cite{zhuo2025streaming} to strengthen its historical context memory efficiency, RegVGGT achieves state-of-the-art performance on long-horizon benchmarks~\cite{palazzolo2019refusion,dai2017scannet,sturm2012benchmark} across diverse FFRM prediction tasks, surpassing prior FFRM-based stream reconstruction baselines~\cite{mahdi2025evict3r,su2026xstreamvggt,yuan2026infinitevggt} comprehensively by a large margin. 
The experimental results showcase the effectiveness of our method, providing valuable insights to the community for adapting FFRMs to stream reconstruction. 

To summarize our contributions:
\begin{enumerate}[topsep=0pt]
    \item We report a heuristic behavior of FFRM inference on image stream input, that the initial saliency of a FFRM token reliably dictates its long-term importance across the stream. 
    \item We propose RegVGGT, which aggressively retains at most 1\% of tokens for each incoming frame from the stream input with minimal compromise in the integrity of historical stream context. 
    \item Extensive experiments show that RegVGGT achieves state-of-the-art performance on long-horizon benchmarks, surpassing baseline methods comprehensively by a large margin. 
\end{enumerate}

\section{Related Works}
\label{sec:related_works}

\subsubsection{Feed-Forward 3D Reconstruction.}
Building upon decades of foundational research~\cite{wu2013towards,furukawa2009accurate,schonberger2016structure}, recent advances have achieved an end-to-end reconstruction paradigm to predict dense 3D geometry directly from input images in a feed-forward manner.
Pioneering feed-forward reconstruction models (FFRMs) such as DUSt3R~\cite{wang2024dust3r} and MASt3R~\cite{leroy2024grounding} 
formulate 3D reconstruction as pairwise pointmap regression, predicting dense 3D geometry by leveraging cross-attention between image pairs, but require costly global alignment post-processing. 
To address this, VGGT~\cite{wang2025vggt} employs global self-attention across all image tokens, predicting point maps and camera poses in a single forward pass, eliminating the need for such alignment.
Subsequent works optimize VGGT across multiple dimensions, such as flexible inputs~\cite{keetha2025mapanything} and equivariant designs~\cite{wang2025pi}.
Despite their impressive reconstruction accuracy, these FFRMs operate in an offline inference regime that forces reprocessing all historical frames whenever a new frame arrives, making it infeasible to process lengthy video streams. 

\subsubsection{Feed-Forward Streaming 3D Reconstruction.}
Streaming 3D reconstruction targets the incremental processing of image streams.
To adapt FFRMs to this paradigm, the critical first step is to avoid reprocessing all historical frames whenever a new frame arrives.
Recent streaming reconstruction methods~\cite{lan2025stream3r,zhuo2025streaming} replace canonical global attention with a temporal causal approach, explicitly retaining all historical Key--Value (KV) pairs as context memory.
This largely alleviates the computational overhead for stream reconstruction, because historical tokens can be directly accessed from memory, eliminating the need to recompute them upon the arrival of each new frame.
However, their unconstrained retention of full KV pairs causes memory usage to grow rapidly, rendering them unsustainable for long-horizon reconstruction.

\subsubsection{Memory Bank Strategy.}
To further extend the capacity of historical frame context embedded into limited memory, various memory bank strategies have been introduced to implicitly or explicitly maintain historical frame tokens as context for reconstructing the incoming frames.
Implicit strategies store latent features from historical frames~\cite{wang20243d,wu2025point3r} or compress historical information into compact hidden states~\cite{wang2025continuous,li2025wint3r,chen2025ttt3r}.
While memory-efficient, they suffer from catastrophic forgetting of crucial historical context as the processed frames increase.
Conversely, recent explicit strategies~\cite{mahdi2025evict3r,su2026xstreamvggt,yuan2026infinitevggt} cap the memory usage by enforcing a fixed memory budget.
However, they rely on the implicit prior of scene scales to properly measure the pre-defined budget.
A fixed budget is inevitably redundant for small-scale scenes and insufficient for large ones, suffering from the same contextual forgetting as the implicit strategies.
In contrast, we admit at most 1\% of salient tokens per frame to update the context memory, dramatically suppressing the memory inflation as the stream progresses with negligible compromise to reconstruction quality.  

\section{Method}
\label{sec:method}

\subsection{Preliminaries}
\label{sec:preliminaries}

\subsubsection{Visual Geometry Grounded Transformer (VGGT)~\cite{wang2025vggt}. }
Our method is to adapt a VGGT-like FFRM to lengthy video stream input. 
A VGGT-like FFRM takes an image sequence of $T$ frames as input, 
\begin{equation}
    \label{eq:1}
    \mathcal{I} = \{\mathbf{I}_t\}_{t=1}^T, \quad \mathbf{I}_t \in \mathbb{R}^{H \times W \times 3},
\end{equation}
to directly predict view-wise camera poses, depth maps and point maps in a single forward pass. 
This process is accomplished with visual transformer~\cite{dosovitskiy2020image}, by encoding images into tokens via an encoder~\cite{oquab2023dinov2}, performing reconstruction in the embedding space with alternated frame-attention and global-attention layers, and finally decoding the tokens with respective prediction heads to produce the dense reconstruction results. 
Though such an architectural design renders impressive reconstruction performance, inference with all input views in one shot and the underlying quadratic growth of memory occupancy against view number prohibit VGGT-like FFRMs from processing lengthy video stream input with hundreds of frames.

\subsubsection{Canonical Attention Operation. }
For a better understanding of the follow-ing discussions, we first introduce canonical attention operation in visual transformer~\cite{dosovitskiy2020image}, along with necessary notations. 
Given three token sequences of length $T\cdot N$ as queries, keys and values, which are denoted as $\mathbf{Q}, \mathbf{K}, \mathbf{V}\in\mathbb{R}^{(T\cdot N)\times D}$ respectively, a canonical attention is defined to calculate, 
\begin{equation}
    \label{eq:2}
    \mathbf{O}=\bigoplus_{h=1}^H{\text{Attn}(\mathbf{Q}_{(h)},\mathbf{K}_{(h)})\times \mathbf{V}_{(h)}}, 
    \text{Attn}\left(\mathbf{Q}_{(h)},\mathbf{K}_{(h)}\right)=\text{Softmax}\left( \frac{\mathbf{Q}_{(h)}\times \mathbf{K}_{(h)}^\top}{\sqrt{D/H}} \right), 
\end{equation}
where $N$ denotes the number of tokens of a frame, $\bigoplus$ denotes concatenation operation along feature dimension, $H$ denotes the number of attention heads, $D$ denotes the dimensional size of token features, and $\mathbf{Q}_{(h)}, \mathbf{K}_{(h)}, \mathbf{V}_{(h)} \in \mathbb{R}^{(T\cdot N)\times \frac{D}{H}}$ so that $\mathbf{Q}=\bigoplus_h\mathbf{Q}_{(h)}, \mathbf{K}=\bigoplus_h \mathbf{K}_{(h)}, \mathbf{V}=\bigoplus_h\mathbf{V}_{(h)}$.
In the remainder of this paper, by default we take $H=1$ for brevity, shortening the $(h)$ in subscript unless otherwise specified.

\subsubsection{Adaption to Stream Input via Temporal Causal Attention. }
To adapt VGGT-like FFRMs to stream input, the first step is to avoid reprocessing all historical frames each time when a new frame is passed. 
This is achieved by replacing the canonical global attention operation in VGGT-like architectures with temporal causal attention operation~\cite{zhuo2025streaming,lan2025stream3r}.  
Specifically, temporal causal attention breaks the queries into view-wise splits that $\mathbf{Q}_t\in\mathbb{R}^{N\times D}$ denotes the query tokens for the $t$-th frame, being the same for keys and values. 
Further notating $\mathbf{K}_{\le t}=\bigcup_{m=1}^{m\le t}{\mathbf{K}_m}$ (the same for $\mathbf{Q}$ and $\mathbf{V}$), there is, 
\begin{equation}
\label{eq:3}
    \mathbf{O}_t=\mathrm{Attn}(\mathbf{Q}_t, \mathbf{K}_{\le t})\times \mathbf{V}_{\le t}, \mathbf{O}_t\in \mathbb{R}^{N\times D}.
\end{equation}
By storing $\mathbf{K}_{\le t-1}$ and $\mathbf{V}_{\le t-1}$ in memory, namely KV caches, one only need to focus on resolving tokens from the $t$-th frame to reconstruct this new incoming frame by efficiently looking up memory for historical context. 
However, retaining complete KV caches in memory lead to a rapid inflation of memory footprint. 

\begin{figure}[t]
  \centering
  \includegraphics[width=\linewidth]{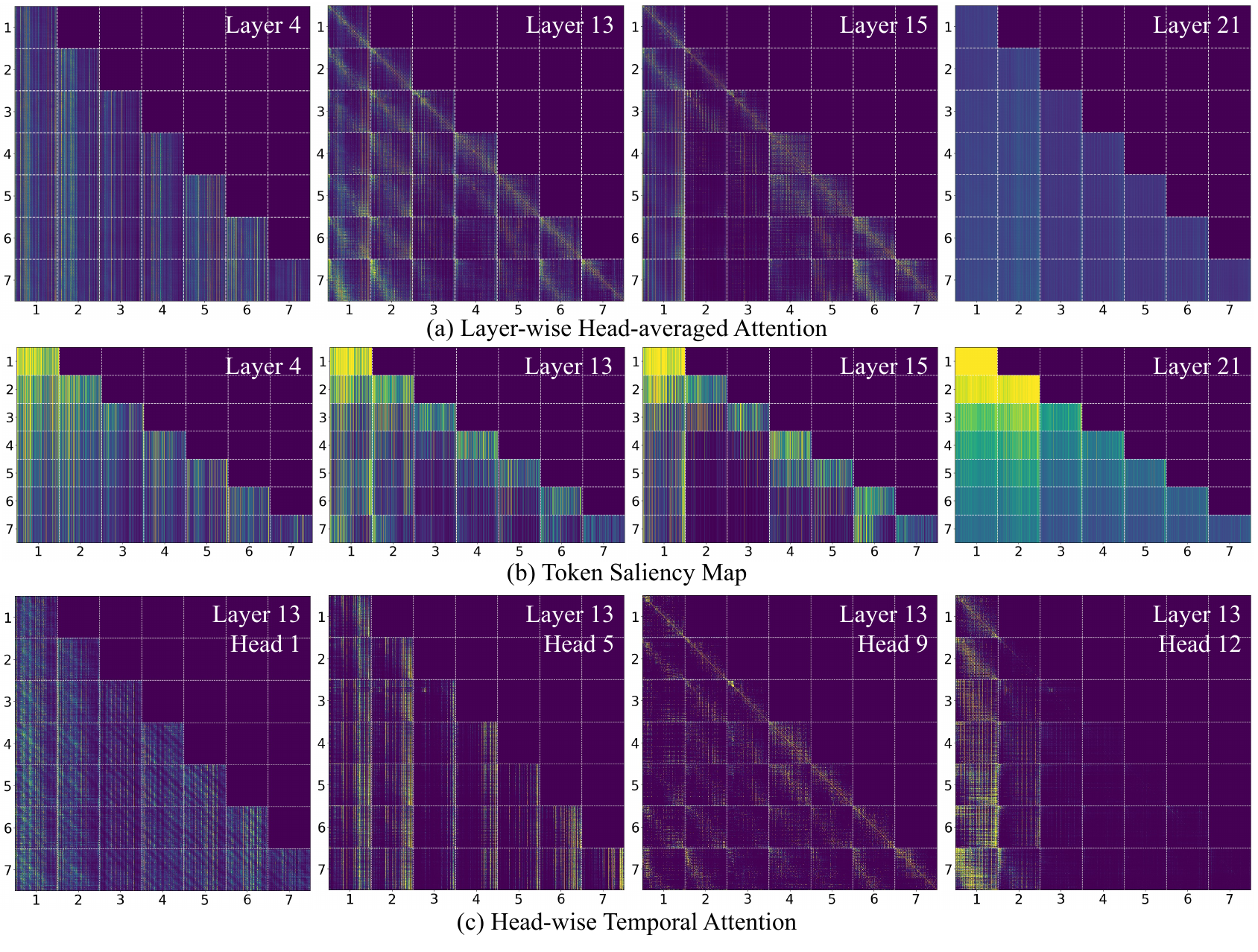}
  \caption{\textbf{Visualization of temporal causal attention.} 
  We visualize attention maps to reveal the interaction between queries (rows) and keys (columns).
  Frames are sampled in a stride of 100 from ScanNet~\cite{dai2017scannet} dataset to depict the long-term behaviour of causal temporal attention. 
  a) We depict layer-wise head-averaged attention map, visualizing $\mathrm{Attn}(\mathrm{Q}_t, \mathrm{K}_{t\le 1})$ in \cref{eq:3} across frame index $t$. 
  b) We sum up each column, \ie the attention of different queries for the same key, showing a temporal non-increasing response of key token saliency. 
  c) The head-wise attention maps within Layer 13. 
  }
  \label{fig:observation}
\end{figure}

\subsection{Insights to FFRMs with Temporal Causal Attention}
\label{sec:motivation}

To mitigate the rapid inflation of memory consumption in temporal causal attention based FFRM, recent and concurrent works~\cite{mahdi2025evict3r,su2026xstreamvggt,yuan2026infinitevggt} artificially adopt a fixed token budget to bound the memory footprint. 
However, as the number of processed frames increases, the budget must decide which memory to abandon for incoming frames, resulting in a distortion of historical frame context. 
Furthermore, it remains unclear \textit{how to ensure that the tokens currently abandoned are not salient for the future frames}. 
We resolve these two questions with three empirical observations, which are discussed below. 

\subsubsection{Observation 1: Temporal Consistency of Token Saliency.}
\label{sec:obs_temporal}
As illustrated in \cref{fig:observation}, the attention scores of each key token are highly consistent with respect to queries from subsequent frames. 
Key tokens assigned high attention scores in their source frame consistently remain salient for the future frames, whereas those with initially low attention scores rarely become salient later.

\noindent\textbf{Implication.} The temporal consistency behaviour of token saliencies indicates that they could be confidently estimated online using attention scores of the current frame, with negligible risk of discarding tokens which are possibly salient for future frames.

\subsubsection{Observation 2: Differentiated Attention Behaviour Across Layers.} 
\label{sec:obs_layer}
We profile the difference in the behaviour of attention maps across different transformer layers, visualizing the head-averaged temporal causal attention maps in \cref{fig:observation} (a). 
In most early and late layers (\eg, Layer 4 and 21, denoted as $\mathcal{L}_\text{qa}$), the attention maps exhibit query-agnostic intensity along the column. 
In contrast, the attention maps in the intermediate layers (\eg, Layers 13 and 15, denoted as $\mathcal{L}_\text{qs}$) display query-specific, sparse and diagonally structured attention patterns. 

\noindent\textbf{Implication.} 
For layers with query-agnostic attention maps, we empirically find that we can only preserve the key/value tokens of the first frame without hindering reconstruction, which is to say zero update in memory is demanded beyond memorizing the first frame. 
It suggests that these high response key tokens might be consistent anchors in physical world to align the subsequent frames with the first frame, thus a preservation of first frame tokens is sufficient to make these layers function well. 
For other layers, their sparse attention intensity suggests a possibility of applying aggressive token pruning on them to suppress the update to historical token memory, minimizing the memory footprint. 

\subsubsection{Observation 3: Distinct Function of Attention Heads.}
\label{sec:obs_head}
Zooming into layers with query-specific attentions, we investigate attention maps at the level of attention head, as shown in \cref{fig:observation}(c). 
The attention maps show head-distinct behaviour, indicating their difference between each other. 

\noindent\textbf{Implication.} 
The saliency of tokens should be evaluated independently for each attention head for best quantification. 

\begin{figure}[t]
  \centering
  \includegraphics[width=\linewidth]{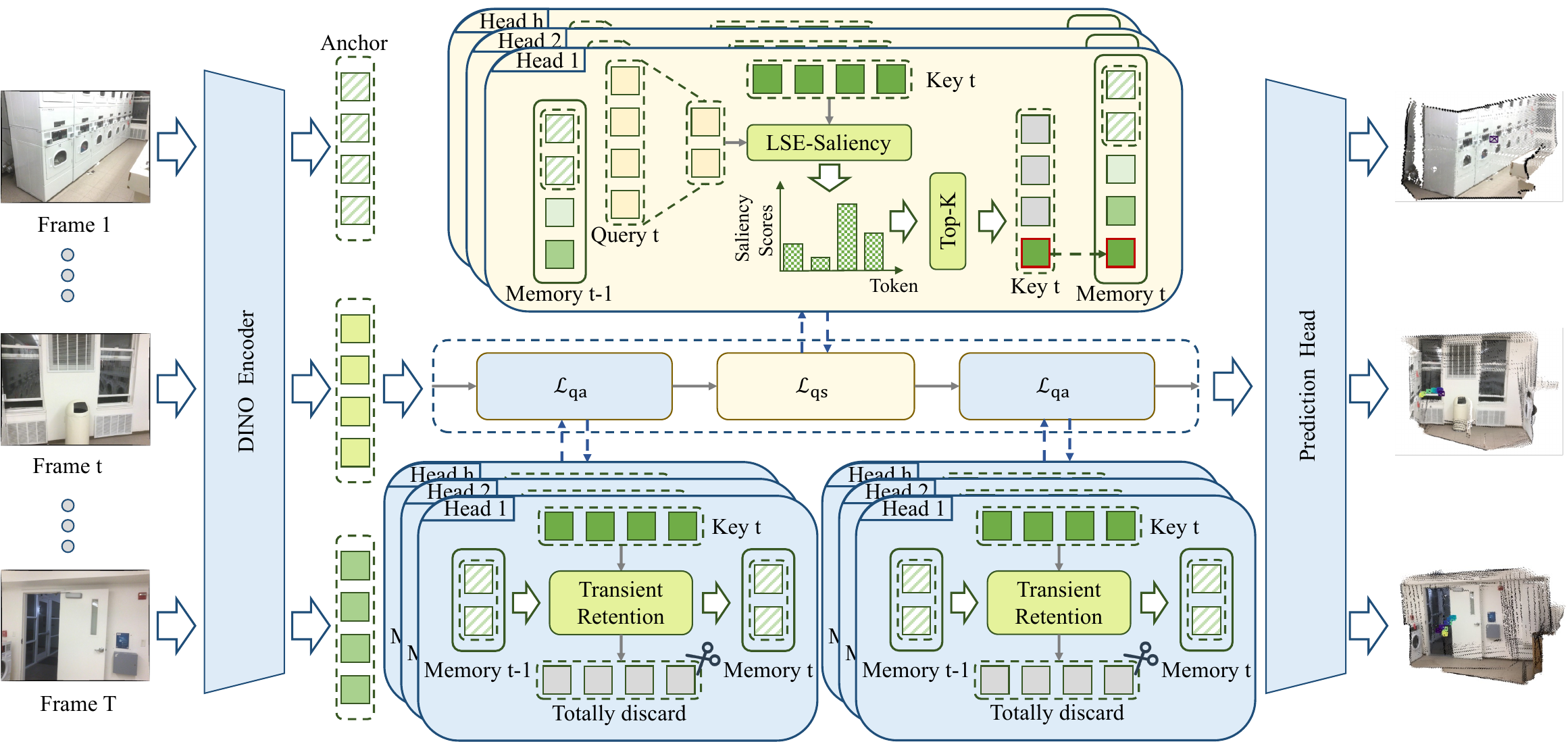}
  \caption{\textbf{Overview of RegVGGT.} Instead of enforcing a fixed memory budget, RegVGGT regulates the per-frame information admitted into memory through a training-free token regulation strategy. 
  We decouple layers into two types. 
  $\mathcal{L}_\text{qa}$ entirely discard intermediate history, while $\mathcal{L}_\text{qs}$ strictly preserve the top 1\% of salient tokens per frame. 
  Coupled with LSE-Saliency, RegVGGT achieves long-term geometric consistency and true scene-agnostic scalability.}
  \label{fig:pipeline}
\end{figure}

\subsection{Proposed Method: RegVGGT}
\label{sec:method_RegVGGT}

As established in Observation 1, token saliency can be reliably estimated on-the-fly using current-frame attention scores, ensuring the most salient tokens retained at the current frame contain the most important context for the future frames. 
Instead of rigidly capping memory footprint, we strictly regulate the information admitted into memory at a constant rate per frame, where the rate can be extremely small according to Observation 2. 
A head-specific token retention strategy is further applied according to Observation 3. 
All proposals above, integrated with a FlashAttention-compatible saliency estimation strategy, together compose our RegVGGT, as illustrated in \cref{fig:pipeline}, which retains at most 1\% tokens per frame to update the historical context memory without compromising reconstruction accuracy.

\subsubsection{Global Anchor Preservation.}
In VGGT-like FFRMs~\cite{wang2025vggt,zhuo2025streaming,lan2025stream3r}, the first frame establishes the global coordinate system to which all subsequent frames are aligned. 
Validating this structural reliance, our empirical analysis reveals that tokens from the first frame receive sustained high attention over time, especially within the intermediate layers (\eg, Layers 13 and 15 in \cref{fig:observation}). 
Pruning these tokens would severely damage long-term geometric consistency. 
Therefore, we designate the complete key--value pairs of the first frame as global anchor tokens, which are permanently preserved in the memory across all $L$ layers: 
\begin{equation}
\label{eq:4}
    \mathcal{M}_{1}^{(l)} = (\mathbf{K}_1^{(l)}, \mathbf{V}_1^{(l)}), \quad \forall l \in \{1, \dots, L\},
\end{equation}
where $\mathbf{K}_1^{(l)}$ and $\mathbf{V}_1^{(l)}$ denote the key/value tokens in the $l$-th layer of the 1-st frame, and $\mathcal{M}_{1}^{(l)}$ represents the corresponding anchor memory.

\subsubsection{Layer-Specific Memory Allocation.}
Motivated by the varying attention behaviour across layers identified in Observation 2, we design a differentiated layer-specific memory allocation strategy for $\mathcal{L}_\text{qa}$ and $\mathcal{L}_\text{qs}$ (see Observation 2 for these two notations). 
For $\mathcal{L}_\text{qa}$, we only preserve the global anchor tokens in the first frame, while the KV caches from all subsequent frames are prohibited from memory, 
\begin{equation}
\label{eq:5}
    \mathcal{M}_t^{(l)} = \mathcal{M}_{\text{anchor}}^{(l)}, \quad \forall l \in \mathcal{L}_{\text{qa}},
\end{equation}
where $\mathcal{M}_t^{(l)}$ denotes the historical memory maintained at temporal index $t$.

In contrast, capitalizing on the diagonally structured attention patterns, we regulate the per-frame information admitted into memory for $\mathcal{L}_\text{qs}$. 
In addition to preserving the global anchor tokens, we retain only the top $k$ most salient tokens per incoming frame, where $k = \lfloor \gamma N \rfloor$,
\begin{equation}
\label{eq:6}
    \mathcal{M}_t^{(l)} =  \mathcal{M}_{t-1}^{(l)} \bigcup \text{Top-}k(\mathbf{K}_t^{(l)}, \mathbf{V}_t^{(l)}) , \quad \forall l \in \mathcal{L}_{\text{qs}},
\end{equation}
where $\gamma \in (0, 1]$ represents the per-frame token retention rate. 
Unlike fixed-budget baselines that enforce a rigid capacity constraint ($|\mathcal{M}_t| \le B$), our memory footprint increases linearly at an extremely low rate, that $|\mathcal{M}_t| = |\mathcal{M}_{1}| + (t-1) \cdot \lfloor \gamma N \rfloor$. 
By setting the retention rate $\gamma$ to merely 0.01, we ensure our memory footprint growth remains highly sustainable.

\subsubsection{Head-Specific Token Retention.}
Motivated by Observation 3, we introduce a head-specific token retention strategy. 
First, to prevent the destructive interference caused by naively averaging attention scores across heads, we evaluate token saliency and maintain memory independently across attention heads within $\mathcal{L}_\text{qs}$.
Second, to minimize computational overhead introduced by saliency estimation, we exploit the inherent spatial redundancy among neighboring queries. 
Specifically, we estimate token saliency using a spatially downsampled subset of queries, denoted as $\tilde{\mathbf{Q}}_t$. 
This subset is obtained by partitioning the current frame's queries into $2 \times 2$ non-overlapping windows and sampling the top-left query from each window. 
The saliency score $S_j$ for the $j$-th token $\mathbf{k}_{t,j}\in\mathbb{R}^{1\times \frac{D}{H}}$ of $\mathbf{K}_t^{(l)}$ in head $h$ is then defined as the cumulative attention it receives from this sampled query subset, which is the sum of the $j$-th column in $\text{Attn}(\tilde{\mathbf{Q}}_t, \mathbf{K}_t) \in \mathbb{R}^{\frac{N}{4}\times N}$, as depicted in the diagonal view patches of \cref{fig:observation}(b). 
Finally, the top-$k$ tokens with the highest saliency scores are selected and retained independently for each attention head, effectively preserving sparse, fine-grained geometric cues.

\subsubsection{FlashAttention-Compatible Saliency Estimation.}
Attention-based token retention strategies, including ours, rely on exact attention scores, which are not explicitly materialized by FlashAttention~\cite{dao2022flashattention,dao2023flashattention}. 
Constructing full attention matrices would incur $\mathcal{O}(N |\mathcal{M}_{\le t}|)$ memory overhead, defeating the purpose of memory efficiency and low latency. 
To address this problem, we introduce LSE-Saliency, a FlashAttention-compatible saliency estimation mechanism that recovers exact attention scores on-the-fly using Log-Sum-Exp (LSE) statistics without materializing full attention maps. 
FlashAttention computes and caches a normalization factor (the log-partition function) for each query during its forward pass:
\begin{equation}
    \text{LSE}_i = \log \sum_{m=1}^t\sum_{j=1}^N \exp(\mathbf{q}_{t,i} \mathbf{k}_{m,j}^\top / \sqrt{\frac{D}{H}}).
\end{equation}
Leveraging this cached statistic, we can efficiently compute the exact attention scores of any token with respect to the entire historical context. 
Given the scaled attention logits between queries and current-frame keys, defined as $Z_{ij} = \mathbf{q}_i \mathbf{k}_j^\top / \sqrt{\frac{D}{H}}$, the exact attention scores can be recovered as:
\begin{equation}
    A_{ij} = \exp(Z_{ij} - \text{LSE}_i), 
\end{equation}
where $A_{i,j}$ denotes the value in the $i$-th row and $j$-th column of $\text{Attn}(\tilde{\mathbf{Q}}_t, \mathbf{K}_t)$.
This allows us to perform fine-grained token retention while fully preserving the hardware efficiency of FlashAttention.

\section{Experiment}
\subsection{Experiment Setup}

We evaluate RegVGGT on multiple 3D tasks, including camera pose estimation, video depth estimation, and 3D reconstruction. 
To assess scalability under long input streams, we focus on test scenes containing a large number of frames.

\subsubsection{Implementation Details.}
RegVGGT is designed as a training-free token regulation strategy. 
We implement it on top of StreamVGGT~\cite{zhuo2025streaming}, keeping the original network architecture and pre-trained weights frozen; our modifications are confined solely to the online token regulation mechanism. 
Furthermore, we adopt the VRAM-efficient implementation from FastVGGT~\cite{shen2025fastvggt}, which discards intermediate features from 20 layers that are not required by subsequent modules. 
Models utilizing this baseline optimization are denoted with an asterisk (\eg, StreamVGGT*). 
Unless otherwise specified, all experiments are conducted under this setting. 
All evaluations are performed on a single NVIDIA RTX 3090 GPU (24GB VRAM).

\subsubsection{Configuration of RegVGGT.}
Guided by the empirical analyses in \cref{sec:motivation}, we partition the 24 Transformer layers into two distinct functional groups. 
We designate Layers 1–10 and 18–24 as $\mathcal{L}_\text{qa}$, and Layers 11–17 as $\mathcal{L}_\text{qs}$. 
For $\mathcal{L}_\text{qs}$, we apply our token regulation strategy by performing a $2 \times 2$ spatial downsampling on queries and setting the retention rate to $\gamma=0.01$ (retaining only 1\% of salient tokens per incoming frame). 
To demonstrate the robustness of our method, these hyperparameter settings remain strictly fixed across all experiments and datasets.

\subsubsection{Fair Comparison Protocol.}
Crucially, to ensure a rigorously fair comparison against baselines, we dynamically align the memory footprint. 
For any given input sequence length and dataset, we calculate the total number of tokens retained by RegVGGT, then enforce identical token budget for all baselines accordingly.
This protocol ensures that all performance comparisons are conducted under identical memory footprints, thereby isolating the effectiveness of our token regulation mechanism.

\begin{table*}[tb]
\scriptsize
\centering
\caption{\textbf{Camera Pose Estimation Results} on TUM Dynamics~\cite{sturm2012benchmark} and ScanNet~\cite{dai2017scannet}.}
\label{tab:camera_pose}
\resizebox{\linewidth}{!}{
\setlength{\tabcolsep}{1.5mm}
\begin{tabular}{l ccccccc}
\toprule
\multirow{2}{*}{\textbf{Model}} 
& \multirow{2}{*}{\textbf{Frames}}
& \multicolumn{3}{c}{\textbf{TUM Dynamics}} 
& \multicolumn{3}{c}{\textbf{ScanNet}} \\
\cmidrule(lr){3-5} \cmidrule(lr){6-8}
& ~ 
& ATE (m)$\downarrow$ & RPE trans$\downarrow$ & RPE rot$\downarrow$ 
& ATE (m)$\downarrow$ & RPE trans$\downarrow$ & RPE rot$\downarrow$ \\
\midrule

Evict3R*~\cite{mahdi2025evict3r}
& \multirow{4}{*}{800}
& 0.162 & 0.028 & 1.572 
& 1.045 & 0.100 & 14.837 \\

InfiniteVGGT*~\cite{yuan2026infinitevggt}
& 
& 0.130 & 0.045 & 1.646 
& 1.007 & 0.138 & 12.405 \\

XStreamVGGT*~\cite{su2026xstreamvggt}  
& 
& 0.121 & 0.020 & 1.354 
& 1.000 & 0.089 & 10.225 \\

RegVGGT* (ours)
& 
& \textbf{0.109} & \textbf{0.016} & \textbf{1.288} 
& \textbf{0.949} & \textbf{0.071} & \textbf{6.662} \\

\midrule

Evict3R*~\cite{mahdi2025evict3r}
& \multirow{4}{*}{900}
& 0.160 & 0.027 & 1.632 
& 1.073 & 0.100 & 15.159 \\

InfiniteVGGT*~\cite{yuan2026infinitevggt}
& 
& 0.132 & 0.043 & 1.709 
& 1.036 & 0.146 & 11.861 \\

XStreamVGGT*~\cite{su2026xstreamvggt}  
& 
& 0.147 & 0.024 & 1.579 
& 1.029 & 0.090 & 10.865 \\

RegVGGT* (ours)
& 
& \textbf{0.125} & \textbf{0.016} & \textbf{1.373} 
& \textbf{0.976} & \textbf{0.072} & \textbf{6.779} \\

\midrule

Evict3R*~\cite{mahdi2025evict3r}
& \multirow{4}{*}{1000}
& 0.174 & 0.027 & 2.004 
& 1.114 & 0.101 & 15.333 \\

InfiniteVGGT*~\cite{yuan2026infinitevggt}
& 
& 0.141 & 0.042 & 1.865 
& 1.061 & 0.141 & 11.622 \\

XStreamVGGT*~\cite{su2026xstreamvggt}  
& 
& 0.174 & 0.027 & 1.866 
& 1.054 & 0.090 & 11.182 \\

RegVGGT* (ours)
& 
& \textbf{0.135} & \textbf{0.016} & \textbf{1.431} 
& \textbf{0.997} & \textbf{0.072} & \textbf{6.656} \\

\bottomrule
\end{tabular}
}
\end{table*}

\subsection{Camera Pose Estimation}
Following CUT3R~\cite{wang2025continuous} and TTT3R~\cite{chen2025ttt3r}, we evaluate camera pose accuracy on the TUM Dynamics~\cite{sturm2012benchmark} and ScanNet~\cite{dai2017scannet} datasets, adopting Absolute Trajectory Error (ATE), Relative Translation Error (RPE trans), and Relative Rotation Error (RPE rot) as our evaluation metrics. 
To systematically assess tracking robustness over long horizons, we follow the protocol established in TTT3R~\cite{chen2025ttt3r}. 
For TUM Dynamics~\cite{sturm2012benchmark}, we extract the initial frames of each sequence. 
For ScanNet~\cite{dai2017scannet}, we apply a temporal stride of 3. 
Both datasets yield evaluation sequences ranging from 50 to 1,000 frames in length.
Quantitative results for extreme sequence lengths of 800, 900, and 1000 frames are presented in \cref{tab:camera_pose}. 
As demonstrated, RegVGGT consistently outperforms all competing methods across all metrics and sequence lengths. 
Notably, as the input length increases, fixed-budget methods experience severe performance degradation, whereas RegVGGT sustains robust tracking accuracy.

\subsection{3D Reconstruction}
Following the standard protocol of CUT3R~\cite{wang2025continuous}, we evaluate 3D reconstruction quality on the 7-Scenes~\cite{shotton2013scene} and NRGBD~\cite{azinovic2022neural} datasets. 
To systematically assess long-horizon performance, we follow TTT3R~\cite{chen2025ttt3r} and sample each sequence with a temporal stride of 2, utilizing the first 50 to 300 frames as input. 
We report performance across three metrics, namely Accuracy (Acc), Completion (Comp), and Normal Consistency (NC). 
Quantitative results for extreme sequence lengths of 200 and 300 frames are presented in \cref{tab:3d_recon}. 
As demonstrated, RegVGGT establishes new state-of-the-art performance across the vast majority of metrics.  
Crucially, as the sequence length scales from 200 to 300 frames, fixed-budget baselines suffer drastic degradation in accuracy and consistency (\eg, Evict3R's mean accuracy error surges by over 50\% on 7-Scenes). 
In contrast, RegVGGT remains highly stable with minimal performance degradation, validating the effectiveness of our method.

\begin{table*}[tb]
\scriptsize
\centering
\caption{\textbf{3D Reconstruction Results} on 7-Scenes~\cite{shotton2013scene} and NRGBD~\cite{azinovic2022neural}.}
\label{tab:3d_recon}
\resizebox{\linewidth}{!}{
\setlength{\tabcolsep}{1mm}
\begin{tabular}{l ccccccccccccc}
\toprule
\multirow{3}{*}{\textbf{Model}} 
& \multirow{3}{*}{\textbf{Frames}}
& \multicolumn{6}{c}{\textbf{7-Scenes}} 
& \multicolumn{6}{c}{\textbf{NRGBD}} \\
\cmidrule(lr){3-8} \cmidrule(lr){9-14}
& ~ 
& \multicolumn{2}{c}{Acc$\downarrow$}
& \multicolumn{2}{c}{Comp$\downarrow$} 
& \multicolumn{2}{c}{NC$\uparrow$}
& \multicolumn{2}{c}{Acc$\downarrow$}
& \multicolumn{2}{c}{Comp$\downarrow$} 
& \multicolumn{2}{c}{NC$\uparrow$}\\
\cmidrule(lr){3-4} \cmidrule(lr){5-6} \cmidrule(lr){7-8} \cmidrule(lr){9-10} \cmidrule(lr){11-12} \cmidrule(lr){13-14}
& ~ 
& Mean & Med. & Mean & Med. & Mean & Med.
& Mean & Med. & Mean & Med. & Mean & Med. \\
\midrule

Evict3R*~\cite{mahdi2025evict3r}
& \multirow{4}{*}{200}
& 0.106 & 0.047 & 0.059 & 0.017 & 0.578 & 0.621 
& 0.107 & 0.061 & 0.035 & 0.010 & 0.638 & 0.741 \\

InfiniteVGGT*~\cite{yuan2026infinitevggt}
& 
& 0.067 & 0.036 & 0.054 & 0.005 & 0.579 & 0.623 
& 0.062 & 0.042 & 0.017 & \textbf{0.004} & 0.631 & 0.725 \\

XStreamVGGT*~\cite{su2026xstreamvggt}  
& 
& 0.075 & 0.038 & 0.029 & 0.005 & 0.581 & 0.627 
& 0.076 & 0.051 & 0.023 & 0.005 & 0.633 & 0.733 \\

RegVGGT* (ours)
& 
& \textbf{0.036} & \textbf{0.020} & \textbf{0.024} & \textbf{0.004} & \textbf{0.591} & \textbf{0.643} 
& \textbf{0.043} & \textbf{0.031} & \textbf{0.016} & \textbf{0.004} & \textbf{0.641} & \textbf{0.745}  \\

\midrule

Evict3R*~\cite{mahdi2025evict3r}
& \multirow{4}{*}{250}
& 0.127 & 0.062 & 0.062 & 0.017 & 0.567 & 0.602 
& 0.117 & 0.072 & 0.050 & 0.013 & 0.633 & 0.729 \\

InfiniteVGGT*~\cite{yuan2026infinitevggt}
& 
& 0.096 & 0.065 & 0.063 & 0.008 & 0.570 & 0.607 
& 0.073 & 0.048 & 0.025 & \textbf{0.004} & 0.628 & 0.720 \\

XStreamVGGT*~\cite{su2026xstreamvggt}  
& 
& 0.102 & 0.054 & 0.030 & 0.007 & 0.571 & 0.610 
& 0.080 & 0.051 & 0.025 & \textbf{0.004} & 0.629 & 0.724 \\

RegVGGT* (ours)
& 
& \textbf{0.037} & \textbf{0.020} & \textbf{0.022} & \textbf{0.004} & \textbf{0.583} & \textbf{0.629} 
& \textbf{0.050} & \textbf{0.030} & \textbf{0.018} & \textbf{0.004} & \textbf{0.640} & \textbf{0.741}  \\

\midrule

Evict3R*~\cite{mahdi2025evict3r}
& \multirow{4}{*}{300}
& 0.163 & 0.091 & 0.080 & 0.032 & 0.552 & 0.579 
& 0.168 & 0.112 & 0.062 & 0.019 & 0.611 & 0.683 \\

InfiniteVGGT*~\cite{yuan2026infinitevggt}
& 
& 0.122 & 0.091 & 0.061 & 0.009 & 0.563 & 0.596 
& 0.081 & 0.058 & \textbf{0.026} & \textbf{0.004} & 0.622 & 0.707 \\

XStreamVGGT*~\cite{su2026xstreamvggt}  
& 
& 0.129 & 0.075 & 0.036 & 0.008 & 0.563 & 0.596 
& 0.102 & 0.068 & 0.035 & 0.005 & 0.618 & 0.702 \\

RegVGGT* (ours)
& 
& \textbf{0.038} & \textbf{0.020} & \textbf{0.022} & \textbf{0.004} & \textbf{0.576} & \textbf{0.617} 
& \textbf{0.068} & \textbf{0.041} & \textbf{0.026} & 0.006 & \textbf{0.630} & \textbf{0.724}  \\
\bottomrule
\end{tabular}
}
\end{table*}

\subsection{Video Depth Estimation}
\begin{figure}[t]
  \centering
  \includegraphics[width=\linewidth]{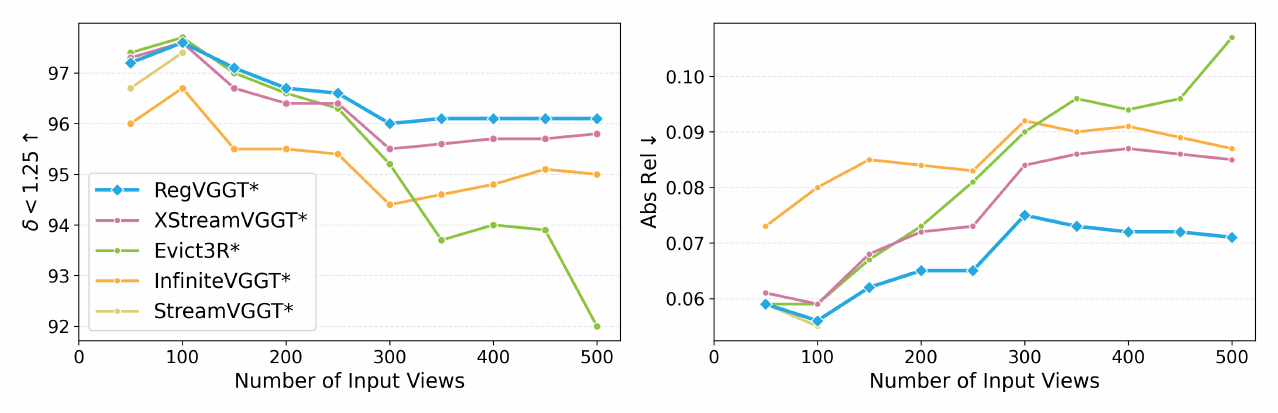}
  \caption{\textbf{Video Depth Estimation Results} on Bonn~\cite{palazzolo2019refusion}.}
  \label{fig:video_depth}
\end{figure}

Since most standard benchmarks contain a highly limited number of consecutive frames, we follow the protocol of TTT3R~\cite{chen2025ttt3r} to evaluate long-horizon performance on the Bonn~\cite{palazzolo2019refusion} dataset. 
Specifically, after discarding the initial 30 frames of each sequence, we sample continuous subsequences ranging from 50 to 500 frames. 
Quantitative results, measured by Absolute Relative error (Abs Rel) and threshold accuracy ($\delta < 1.25$, denoting the percentage of predicted depths within a factor of 1.25 of the ground truth), are presented in \cref{fig:video_depth}.  
While Evict3R~\cite{mahdi2025evict3r} performs competitively with RegVGGT on ultra-short sequences (50 and 100 frames), our model rapidly establishes state-of-the-art dominance as the temporal horizon extends. 
Crucially, as the input length increases, the performance gap between our method and fixed-budget baselines widens substantially. 
This diverging trend directly highlights the superior scalability and robustness of our regulated memory.

\begin{table*}[tb]
\centering
\scriptsize
\renewcommand{\arraystretch}{1.15}
\setlength{\tabcolsep}{1.5mm} 

\begin{minipage}[c]{0.48\linewidth}
\centering
\captionof{table}{\textbf{Ablation Study on Global Anchor Preservation.}}
\label{tab:global_anchor}
\resizebox{\linewidth}{!}{
\begin{tabular}{lccc}
\toprule
\textbf{Method} & ATE (m)$\downarrow$ & RPE trans$\downarrow$ & RPE rot$\downarrow$ \\
\midrule
w/o Global Anchor & 0.175 & 0.063 & 4.005 \\
w/ Global Anchor & \textbf{0.165} & \textbf{0.061} & \textbf{3.930} \\
\bottomrule
\end{tabular}
}
\end{minipage}%
\hfill 
\begin{minipage}[c]{0.48\linewidth}
\centering
\captionof{table}{\textbf{Ablation Study on Head-Specific Token Retention.}}
\label{tab:head_method}
\resizebox{\linewidth}{!}{
\begin{tabular}{lcccc}
\toprule
\textbf{Method} & ATE (m)$\downarrow$ & RPE trans$\downarrow$ & RPE rot$\downarrow$ & Latency (ms)$\downarrow$ \\
\midrule
w/o Head-specific Retention & 0.165 & 0.061 & \textbf{3.840} & 17.66 \\
w/ Head-specific Retention & 0.165 & 0.061 & 3.930 & \textbf{8.29} \\
\bottomrule
\end{tabular}
}
\end{minipage}

\begin{minipage}[c]{0.48\linewidth}
\centering
\captionof{table}{\textbf{Ablation Study on Layer-Specific Memory Allocation.}}
\label{tab:layer_method}
\resizebox{\linewidth}{!}{
\begin{tabular}{lccc}
\toprule
\textbf{Method} & ATE (m)$\downarrow$ & RPE trans$\downarrow$ & RPE rot$\downarrow$ \\
\midrule
w/o Layer-Specific Retention & 0.176 & 0.066 & 4.970 \\
w/ Layer-Specific Retention & \textbf{0.165} & \textbf{0.061} & \textbf{3.930} \\
\bottomrule
\end{tabular}
}
\end{minipage}%
\hfill 
\begin{minipage}[c]{0.48\linewidth}
\centering
\captionof{table}{\textbf{Sensitivity Analysis of Query Downsampling Factor.}}
\label{tab:downsample}
\resizebox{\linewidth}{!}{
\begin{tabular}{lcccc}
\toprule
\textbf{Downsample Factor} & ATE (m)$\downarrow$ & RPE trans$\downarrow$ & RPE rot$\downarrow$ & Latency (ms)$\downarrow$\\
\midrule
$1 \times 1$ & \textbf{0.165} & \textbf{0.060} & \textbf{3.735} & 17.64\\
$2 \times 2$ (default) & \textbf{0.165} & 0.061 & 3.930 & 8.29\\
$3 \times 3$ & 0.167 & 0.062 & 4.224 & 7.69\\
$4 \times 4$ & 0.168 & 0.062 & 4.260 & \textbf{7.45} \\
\bottomrule
\end{tabular}
}
\end{minipage}

\end{table*}

\subsection{Ablation Study}
We conduct ablation studies on the ScanNet~\cite{dai2017scannet} dataset to systematically evaluate the contribution of each key component in our architecture. 
Beyond validating the effectiveness of individual architectural components, we provide an in-depth analysis of critical parametric choices.

\subsubsection{Effectiveness of Core Components.} We ablate the global anchor preservation, the layer-specific memory allocation and the head-specific token retention under an identical total token budget to evaluate their individual contributions. 
As demonstrated in \Cref{tab:global_anchor,tab:layer_method}, independently removing either the global anchor or the layer-specific memory allocation leads to a substantial degradation in tracking accuracy. 
Furthermore, while these two mechanisms secure tracking robustness, \cref{tab:head_method} reveals the critical efficiency contribution of our head-specific token retention strategy. 
Our strategy slashes the inference latency by more than half with a negligible impact on primary accuracy. 
Ultimately, these results confirm that all three components are indispensable to our architecture.

\subsubsection{Sensitivity to Token Retention Rate.} 
\begin{figure*}[tb]
\centering
\scriptsize
\renewcommand{\arraystretch}{1.15}
\setlength{\tabcolsep}{1.5mm} 
\begin{minipage}[c]{0.47\linewidth}
\centering
\includegraphics[width=\linewidth]{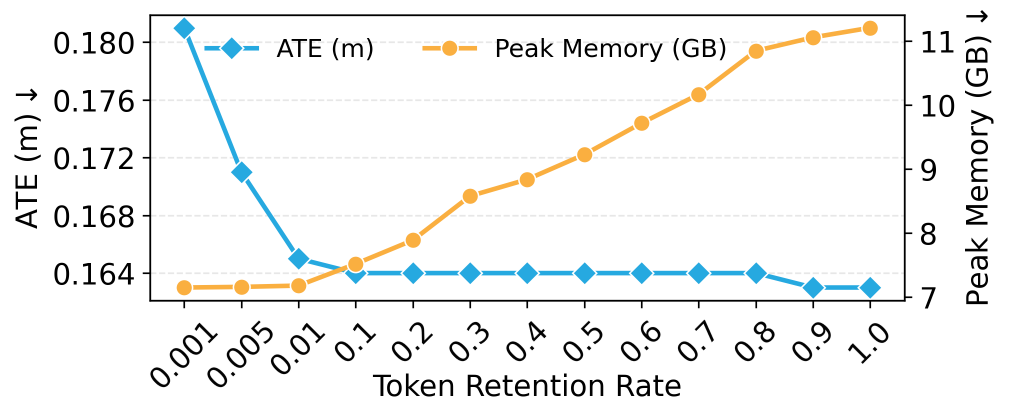}
\caption{\textbf{Sensitivity Analysis of Token Retention Rate.}}
\label{fig:rate_ablation}
\end{minipage}
\hfill 
\begin{minipage}[c]{0.49\linewidth}
\centering
\includegraphics[width=\linewidth]{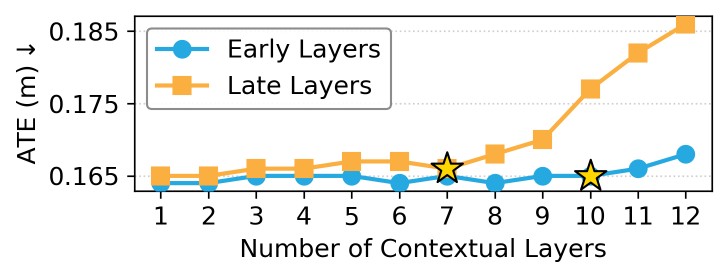}
\caption{\textbf{Sensitivity Analysis of Layer Decoupling.}}
\label{fig:layer_ablation}
\end{minipage}%
\end{figure*}

We analyze the effect of the token retention rate $\gamma$ on tracking accuracy and peak memory footprint. 
As shown in~\cref{fig:rate_ablation}, aggressively reducing $\gamma$ below $0.01$ leads to severe performance deterioration while yielding negligible memory savings, indicating that critical geometric cues are destructively over-pruned. 
However, performance rapidly stabilizes and remains highly robust once $\gamma \geq 0.01$. 
Furthermore, pushing $\gamma$ beyond $0.1$ provides negligible geometric benefits while continuously inflating the peak memory footprint. 
This trade-off curve demonstrates that retaining a mere 1\% of salient tokens per frame is entirely sufficient for RegVGGT to preserve strict geometric consistency. 
Consequently, we adopt $\gamma = 0.01$ as the default configuration across all experiments.

\subsubsection{Sensitivity to Layer Decoupling.} Our method decouples Transformer layers into $\mathcal{L}_\text{qa}$ and $\mathcal{L}_\text{qs}$ based on the differentiated attention behaviour across layers revealed in \cref{sec:motivation}. 
To translate this qualitative observation into an optimal architectural configuration, we evaluate the sensitivity to layer decoupling. 
Specifically, we progressively designate varying proportions of early and late layers as $\mathcal{L}_\text{qa}$, while treating the remaining intermediate layers as $\mathcal{L}_\text{qs}$. 
As shown in \cref{fig:layer_ablation}, we empirically determine the optimal decoupling to be the first 10 and last 7 layers (Layers 1–10 and 18–24) acting as $\mathcal{L}_\text{qa}$, and the intermediate 7 layers (Layers 11–17) acting as $\mathcal{L}_\text{qs}$. 
By discarding the entire intermediate history across 75\% of the layers, this highly asymmetric design strikes a remarkably favorable balance between state-of-the-art accuracy and extreme memory efficiency.

\subsubsection{Sensitivity to Query Downsampling.} We evaluate the sensitivity of the downsampling factor on tracking accuracy and the extra inference latency introduced by saliency estimation. 
As presented in~\cref{tab:downsample}, applying a $2 \times 2$ downsampling factor drastically reduces the inference latency by more than 50\% while achieving near-lossless primary trajectory accuracy. 
However, adopting more aggressive downsampling ($3 \times 3$ and beyond) yields severely diminishing returns in computational speedups while introducing noticeable accuracy degradation. 
Therefore, we adopt the $2 \times 2$ downsampling factor as our default configuration, securing an optimal trade-off between efficiency and tracking accuracy.

\begin{figure*}[tb]
\centering
\scriptsize
\renewcommand{\arraystretch}{1.15}
\setlength{\tabcolsep}{1.5mm} 
\begin{minipage}[c]{0.48\linewidth}
\centering
\includegraphics[width=\linewidth]{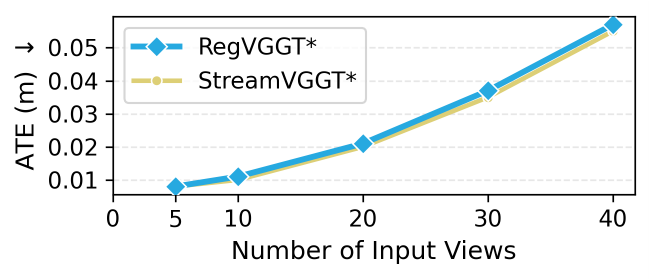}
\caption{\textbf{Analysis of Context Integrity.}}
\label{fig:analysis_ab}
\end{minipage}
\hfill 
\begin{minipage}[c]{0.48\linewidth}
\scriptsize
\centering
\captionof{table}{\textbf{Generalizability evaluation.}}
\label{tab:final}
\resizebox{\linewidth}{!}{
\begin{tabular}{lccc}
\toprule
\textbf{Method} & Frames & Abs Rel$\downarrow$ & $\delta < 1.25 \uparrow$ \\
\midrule
Evict3R*~\cite{mahdi2025evict3r} & \multirow{2}{*}{400} & 0.073 & 95.8\\
Regulated Evict3R*~\cite{mahdi2025evict3r} &  & \textbf{0.071} & \textbf{96}\\
\midrule
Evict3R*~\cite{mahdi2025evict3r} & \multirow{2}{*}{500} & 0.076 & 95.8\\
Regulated Evict3R*~\cite{mahdi2025evict3r} &  & \textbf{0.069} & \textbf{96.1} \\
\bottomrule
\end{tabular}
}
\end{minipage}
\end{figure*}

\subsubsection{Analysis of Context Integrity.}
To validate our claim that RegVGGT achieves extreme memory efficiency with negligible compromise to model performance, we conduct a comparison against the uncompressed FFRM backbone. 
Because the canonical backbone suffers from rapid memory inflation on lengthy streams, we strictly limit the evaluation lengths of this experiment to short sequences ranging from 5 to 40 frames. 
As illustrated in \cref{fig:analysis_ab}, despite retaining at most 1\% of the most salient tokens per frame, RegVGGT exhibits negligible compromise to performance across all evaluation metrics. 
This compelling result confirms that our method successfully isolates the geometrically crucial information, thoroughly demonstrating that the omitted historical tokens are indeed highly redundant for dense 3D reconstruction.

\subsubsection{Generalizability of Token Regulation Strategy.}
To further validate the effectiveness and generalizability of our token regulation strategy, which retains a mere 1\% of salient tokens per frame, we conduct a plug-and-play experiment. 
Existing streaming methods, such as Evict3R~\cite{mahdi2025evict3r}, typically employ a fixed global token budget. However, such a rigid constraint is inevitably redundant for small-scale scenes and insufficient for large ones, inevitably leading to catastrophic forgetting as the input sequence grows. 
To address this, we replace the fixed budget module in Evict3R~\cite{mahdi2025evict3r} with our proposed per-frame regulation strategy, ensuring that the evaluation is conducted under identical token budget. 
The video depth estimation results on the Bonn dataset~\cite{palazzolo2019refusion} are presented in \cref{tab:final}. 
As illustrated, our regulated Evict3R achieves consistent improvements across all metrics. 
This finding not only demonstrates that a dynamic, sequence-length-aware capacity is intrinsically superior to a fixed budget, but also highlights the strong transferability of our design across different baseline architectures.

\subsubsection{Robustness in Challenging Scenarios.}
We simulate highly dynamic environments and large motions via aggressive 1/100 frame sampling. Specifically, we evaluate camera pose estimation on the TUM Dynamics~\cite{sturm2012benchmark} and point map reconstruction on 7-Scenes~\cite{shotton2013scene} (\cref{tab:hard_case}). RegVGGT robustly maintains the backbone's performance in highly dynamic settings. As 7-Scenes results show, large motion sharply reduces temporal redundancy by introducing abundant unique per-frame geometry, greatly alleviated by adopting a milder retention rate of $\gamma = 0.1$.

\subsubsection{Quantitative Metrics for Observation 1.}
Using Layer 13 (\cref{fig:observation}) as an example, we quantify temporal consistency on ScanNet~\cite{dai2017scannet} by computing the Spearman rank correlation of per-token saliency relative to the 10th frame (\cref{tab:spearman_corr}). Heads 1 and 5 maintain robust temporal consistency, while Heads 9 and 12 naturally decay as they prioritize global consistency and the reference frame.

\begin{table*}[tb]
\centering
\scriptsize
\renewcommand{\arraystretch}{1.15}
\setlength{\tabcolsep}{1.5mm} 

\begin{minipage}[c]{0.50\linewidth}
\centering
\captionof{table}{\textbf{Robustness Evaluation in Challenging Scenarios.}}
\label{tab:hard_case}
\resizebox{\linewidth}{!}{
\begin{tabular}{l cccccc}
\toprule
\multirow{2}{*}{\textbf{Model}} 
& \multirow{2}{*}{\textbf{$\gamma$}} 
& \multicolumn{3}{c}{\textbf{Camera Pose}}
& \multicolumn{2}{c}{\textbf{Point Map}}\\
\cmidrule(lr){3-5} \cmidrule(lr){6-7}
& 
& ATE $\downarrow$ & RPE-t $\downarrow$ & RPE-rot $\downarrow$
& CD $\downarrow$ & NC $\uparrow$  \\
\midrule
StreamVGGT* & 1.00 & 0.159 & \textbf{0.668} & \textbf{48.970} & \textbf{0.054} & \textbf{0.750} \\
RegVGGT*    & 0.10 & \textbf{0.154} & 0.673 & 49.010 & 0.066 & 0.745 \\
RegVGGT*    & 0.01 & 0.155 & 0.677 & 49.026 & 0.099 & 0.731 \\
\bottomrule
\end{tabular}
}
\end{minipage}%
\hfill 
\begin{minipage}[c]{0.48\linewidth}
\centering
\captionof{table}{\textbf{Quantitative Metrics for Observation 1.}}
\label{tab:spearman_corr}
\resizebox{\linewidth}{!}{
\begin{tabular}{l ccccc}
\toprule
\multirow{2}{*}{\textbf{Attention Head}} & \multicolumn{5}{c}{\textbf{Time Offset ($\Delta t$)}} \\
\cmidrule(lr){2-6}
& +10 & +30 & +50 & +70 & +90 \\ 
\midrule
Head 1 & 0.997 & 0.990 & 0.980 & 0.979 & 0.976 \\
Head 5 & 0.997 & 0.983 & 0.967 & 0.954 & 0.943 \\
\midrule 
Head 9  & 0.820 & 0.603 & 0.527 & 0.507 & 0.439 \\
Head 12 & 0.926 & 0.739 & 0.541 & 0.479 & 0.299 \\
\bottomrule
\end{tabular}
}
\end{minipage}

\end{table*}

\section{Conclusion}
In this paper, we presented RegVGGT, a training-free online token regulation method that resolves the trade-off between rapid memory inflation and degraded context integrity in FFRMs.
Driven by the key observation that the initial saliency of a FFRM token reliably dictates its long-term importance across the stream, our method aggressively regulates the context by retaining at most 1\% of the most salient tokens per frame.
Coupled with a FlashAttention-compatible saliency estimation mechanism, RegVGGT enables the processing of thousands of frames on consumer-grade GPUs with negligible compromise to reconstruction quality.
Extensive experiments demonstrate that our method comprehensively surpasses existing baselines on long-horizon benchmarks, providing valuable insights to the community for adapting FFRMs to stream reconstruction.

\section*{Acknowledgements}
The research was supported by the National Natural Science Foundation of China (U23B2009, 62471158).

%
%
\bibliographystyle{splncs04}
\bibliography{main}
\end{document}